\documentclass[letterpaper]{article}
\usepackage{aaai}
\usepackage{times}
\usepackage{helvet}
\usepackage{courier}
\usepackage{amssymb}
\usepackage{amsmath}
\usepackage{algorithm}
\usepackage{algorithmic}
\usepackage{setspace}
\usepackage{graphicx}
\usepackage{stfloats}
\usepackage{verbatim}
\usepackage{hyperref}
\usepackage{pgfplots}   

\begin{document}
%
\title{Meta Reinforcement Learning with Distribution of Exploration Parameters Learned by Evolution Strategies}
\author{Yiming Shen\textsuperscript{1}, Kehan Yang\textsuperscript{2}, Yufeng Yuan\textsuperscript{3}, Simon Cheng Liu\textsuperscript{1}\\\\
\textsuperscript{1} LevelupAI
\\
\textsuperscript{2} Beijing University of Posts and Telecommunications\\
\textsuperscript{3} Beijing Normal University}

\maketitle

\begin{abstract}
\begin{quote}
In this paper, we propose a novel meta-learning method in a reinforcement learning setting, based on evolution strategies (ES), exploration in parameter space and deterministic policy gradients. 
ES methods are easy to parallelize, which is desirable for modern training architectures; however, such methods typically require a huge number of samples for effective training. 
We use deterministic policy gradients during adaptation and other techniques to compensate for the sample-efficiency problem while maintaining the inherent scalability of ES methods. We demonstrate that our method achieves good results compared to gradient-based meta-learning in high-dimensional control tasks in the MuJoCo simulator. 
In addition, because of gradient-free methods in the meta-training phase, which do not need information about gradients and policies in adaptation training, we predict and confirm our algorithm performs better in tasks that need multi-step adaptation.
\end{quote}
\end{abstract}

\section{Introduction}
Deep reinforcement learning, which combines deep learning and reinforcement learning, has achieved significant progress recently. 
The performance of state-of-the-art algorithms is close to or even better than human performance in Atari games \cite{Mnih2013Playing}, Go \cite{Silver2016Mastering,Silver2017Mastering} and even multiplayer online games such as Dota. 
However, one inherent drawback of deep reinforcement learning is the tendency to overfit to the current environment setting, 
which makes agents unable to adapt quickly to slight variations in the environment. 

Approaches combining deep reinforcement learning and meta learning are proposed by researchers to address this problem, 
with the aim of improving the applicability of deep reinforcement learning to real-world problems. A store of prior knowledge, or ``common sense", is very important for humans learning to perform a new task quickly, and we can use experience from previous tasks and fast adaptations to perform a new, similar task, integrating this prior knowledge into the initial parameters of neural networks. One common scheme for this approach is to learn a good initial parameter configuration integrating common knowledge of a distribution of tasks; the agent can then quickly find the appropriate parameters if a particular task is in the distribution. 

Recent research following this scheme, such as Model-Agnostic Meta-Learning (MAML) \cite{Finn2017Model}, 
mainly focus on gradient-based methods; these methods achieve state-of-the-art results in multi-task environments.
However, gradient-based methods need higher-order gradients to train initial parameters; in reinforcement learning, these methods use Trust Region Policy Optimization (TRPO) to improve stability by limiting changes between the initial and adapted policies. However, this constraint is restrictive when many steps are needed to train an adapted policy during. (updated)For example, MAML will crash if it takes several steps gradients updates during adapt-training in training stage, which will be showed in the following experiments.

To resolve these problems of gradient-based meta-learning, we consider the methods that do not need gradients in adaptation training, like evolution strategies (ES), which search the distribution of parameters for 
neural networks.
However, meta-learning with evolution strategies is less data-efficient than gradient-based meta-learning; (update)the non-gradient zero order optimizer of evolution strategies may need to sample more data when performing adaptive training than MAML does, and is difficult to train on tasks that have continuous actions and continuous distributions. 

It is important to improve data efficiency in adaptation training and make use of parameter noise in meta-training. We consider that we can use a deterministic policy gradient algorithm to improve data sampling efficiency during adapt-training stage, and use the noise from evolution strategies to explore varied strategies during meta-training stage(update), because the Gaussian noise in parameters for a neural network can be used to try out new strategies in reinforcement learning\cite{parameters_space_noise}; using more noise in meta-training can therefore improve training speed of evolution strategies.

We propose a novel approach, combining evolution strategies, parameter space noise and deterministic policy gradients to tackle the problem of meta-learning in a reinforcement learning setting. 
The key idea behind our approach is to enable the agent to learn the shared prior knowledge of a collection of tasks while exploring and sampling efficiently. 
The agent is represented by a meta-distribution of policies which, in fact, are Gaussian distributions over each parameter learned by evolution strategies. 
The mean values of parameters represent an overall good initial policy on the whole collection of tasks, while the standard deviations of parameters indicate how much such parameters should be tweaked to adapt to a specific new task. 
(xiugai)By using the different combinations of sampled policies according to the learned standard deviation, the sample-efficiency and training time of evolution strategies can be improved using our approach. 
 (update)The meta policy we are learning is a distribution instead of a deterministic policy, and we are learning meta policy and exploration strategies together, embedded in the meta-distribution, instead of learning them separately. One advantage of our approach is that it is easy to deploy in a parallel framework, since it does not need to compute gradients across parallel workers, and the performance grows almost linearly as the number of parallel workers is increased, without much cost due to increased communication between workers.

We apply our algorithms to several benchmark \cite{Duan2016Benchmarking} problems in Mujoco environments, such as half-Cheetah and Ant with random target speeds and random goals. And the results show the performance of our methods is close to or even better than the methods in MAML \cite{Finn2017Model}.

\section{Related Work}

\subsection{Evolution Strategies}
The method of evolution strategies is inspired by the process of natural evolution \cite{Back1991A}. 
The basic idea behind ES is as follows: a population, represented by policy parameters, 
is slightly perturbed at every generation to generate multiple new children. 
The performance of each child will be evaluated by a fitness function, which is an indicator of the benefit of 
the perturbation exerted on the population. The beneficial perturbations will be kept and reused in later generations. 
This iterative procedure will be repeated until a good solution for the objective is found. Current ES methods follow the above scheme and differ primarily in specific methods used for perturbation and selection. As a black-box optimization method, it has several desirable properties compared with the gradient-based methods more widely used in reinforcement learning today \cite{Salimans2017Evolution}: 
\begin{itemize}
\item No need to compute the gradients and back-propagate them.
\item Well adapted to environments whose rewards are sparsely-distributed.
\item Computation can be easily scaled to multiple parallel workers.
\item Indifferent to arbitrary length of horizon.
\end{itemize}
The specific ES method we use in our work belongs to the class of natural evolution strategies \cite{Schaul2008Natural}, which maintains a search distribution for perturbation and iteratively updates the distribution using the estimated gradients with respect to the fitness function. The general procedure of NES can be described as follows: In every iteration, the parameterized search distribution generates a batch of search points and a fitness function will be used to evaluate the performance of every point. Then, the gradients with respect to the fitness function will be computed to update the search distribution, in order to maximize the expected score on the current distribution.
If we use $\theta$ to denote the parameters of probability density of search distribution $\pi(z|\theta)$ and $f(z)$ to denote the fitness function for sample $z$, the expected search gradient and the estimate of the search gradient from samples $z_1...z_\lambda$ can be written as
\begin{equation}
\begin{aligned}
\nabla_{\theta}J(\theta)  & =E_\theta[f(z)\nabla_\theta \log\pi(z|\theta)] \\
 & \approx \frac{1}{\lambda}\sum^\lambda_{k=1} f(z_k)\nabla_\theta \log \pi(z_k|\theta)
\end{aligned}
\end{equation}

\subsection{Parameter Space Noise}

Efficient and consistent exploration can prevent agents from converging prematurely on a local optimum and allow them to continue searching for a better one, which is crucial in gaining better performance. This is even more important in the meta-learning setting, since the agent needs to explore to understand the current environment. Various exploration methods have been proposed to address this problem. In deterministic methods, $\epsilon$-greedy exploration, softmax exploration and UCB exploration \cite{Pecka2014Safe} have been proposed, while in stochastic methods, the policy itself is a distribution over actions. However, most of the methods today focus on noise in action space, which might result in discarding all temporal structure and gradient information. 

When Gaussian action noise is used, the action is sampled from a stochastic policy $a \sim \pi_\theta(s_t) + \mathcal{N}(0, \sigma^2I)$ in which the stochasticity is independent of the current state. Therefore, even two identical states sampled in rollouts might result in the agent choosing completely different actions. If we denote the parameters of our model as $\theta$, the basic idea for parameter space noise \cite{Plappert2018Parameter} is to use policy $\theta^{'}=\theta + \mathcal{N}(0, \sigma^2I)$ for more consistent exploration. The perturbed policy $\theta^{'}$ is sampled at the beginning of a rollout and kept fixed during the entire trajectory. For off-policy methods, the parameter space noise can be directly applied to parameters, since data is collected offline in such methods. In this case, the perturbed policy $\theta^{'}$ is used to collect samples while the non-perturbed policy $\theta$ will be trained. 

\subsection{Deterministic Policy Gradient}
Policy gradient algorithms \cite{Sutton1999Policy} are widely-used in reinforcement learning. The basic idea behind them is to represent the policy as a parametric probability density distribution $\pi(a|\theta)=P[a|s;\theta]$, then the action actually taken can be sampled from such distributions. To train the policy is basically to move the distribution in the direction of higher reward using its gradient. The counterpart of stochastic policy gradients is deterministic policy gradients \cite{Mnih2015Human}, which represent the policy $a = \mu_\theta(s)$ as the mapping from the current state to an exact action instead of a distribution. In the stochastic policy gradient theorem, the policy gradient can be written as:
\begin{equation}
\begin{aligned}
\nabla_\theta J_\theta(\pi_\theta) & = \int_{\mathcal{S}} \rho^\pi(s)\int_{\mathcal{A}}\nabla_\theta\pi_\theta(a|\theta)Q^\pi(s,a)dads \\
& = E_{s\sim \rho^\pi, a\sim \pi_\theta} [\nabla_\theta\log\pi_\theta(a|s)Q^\pi(s,a)]
\label{xxx}
\end{aligned}
\end{equation}
The stochastic policy gradient needs to integrate over both action space and state space, so more samples are required, especially in high-dimensional action space. However, the deterministic policy gradient only integrates over the state space: 
\begin{equation}
\begin{aligned}
\nabla_\theta J_\theta(\mu_\theta) & = \int_{\mathcal{S}} \rho^\mu(s)\nabla_\theta\mu_\theta(s)\nabla_a Q^\mu(s,a)|_{a=\mu_\theta(s)} ds \\
& = E_{s\sim \rho^\mu} [\nabla_\theta\mu_\theta(s)\nabla_a Q^\mu(s,a)]
\label{xxx}
\end{aligned}
\end{equation}
This simpler form means that the deterministic policy gradient can be estimated much more efficiently than the usual stochastic version.

\section{Meta Reinforcement Learning with Distribution of Exploration Parameters Learned by Evolution Strategies}

\subsection{Meta Reinforcement Learning Problem Setup}
In meta-learning for RL, each task $\mathcal{T}_i$ consists of an initial state distribution $p_i(s_1)$, a transition distribution $p_i(s_{t+1}|s_t, a_t)$ and a loss function $\mathcal{L}_{\mathcal{T}_i}$ corresponding to the reward function $R_i$. If we denote the length of the horizon in such a Markov decision process as $H$ and the model of the agent as $\theta$, then the loss for task $\mathcal{T}_i$ and model $\theta$ takes the form:
\begin{equation}
\mathcal{L}_{\mathcal{T}_i}(\theta) = E_{s_t, a_t \sim \theta, \mathcal{T}_i} \Bigg[\sum^{H}_{t=1} R_i(s_t, a_t) \Bigg]
\end{equation}
If we denote $\mathcal{T}$ as the collection of tasks $\mathcal{T}_i$ and $p(\mathcal{T})$ as the distribution of $\mathcal{T}$, the overall loss on all tasks can be written as follows:
\begin{equation}
\mathcal{L}(\theta) = E_{\mathcal{T}_i \sim p(\mathcal{T})}\big[\mathcal{L}_{\mathcal{T}_i}(\theta)\big]
\end{equation}
In $K$-shot learning, which we are focusing on, $K$ rollouts can be acquired from the current policy and model. Those rollouts will be used for adaptation training on the current task and the performance of the agent will be evaluated after adaptation training.

\subsection{Outline of our Algorithms}
In our algorithm, we use zero order optimization of evolution strategy\cite{Salimans2017Evolution} to train the meta model and DDPG\cite{DDPG} to do adapting training. In every iteration, we use meta model which is a policy distribution $\mathcal{N}(\mu,\,\sigma^{2})$ to get M workers. After adapt training the M workers parallel, the scores of rollouts by these workers used as fitness to train the meta model by evolution strategy.

The algorithms we propose can be viewed as an outer loop for meta-training and an inner loop for adaptation training. 
In the adaptation training phase, the agent will be trained on the same tasks for a few iterations while in the meta-training phase, 
the environment will be switched to different tasks.

The i-th worker will be adaptively trained by the following ways. In the beginning of adaptation training, $K$ actor models' parameters $\theta^i_{1}$,$\theta^i_{2}$...$\theta^i_{K}$ and a critic model's parameters $\phi^i$ will be sampled from the policy distribution $P_{actor}(\theta|\mu_a,\sigma_a)$ and $P_{critic}(\phi|\mu_c,\sigma_c)$. These two probability distributions are meta model learning in meta-training to explore and sample in current tasks.
The actor model whose parameters are the means of the $K$ sampled actor models' and the sampled critic model parameters will be used as the initial actor's and critic's parameters for adaptation training. 
In addition, the actor models $\theta^i_{1}$,$\theta^i_{2}$...$\theta^i_{K}$ , which can be regarded as $\overline{\theta}^i$ with noise in actor parameter space, will be used to sample in current tasks for $\overline{\theta}^i$ to train by DDPG.

The trained actor parameters $\overline{\theta^i}'$  and critic parameters $\phi'$ will serve as a worker to rollout to get the corresponding performance used in meta training by evolution strategy.
In one meta iteration, the sampling and adapt-training process of $\overline{\theta}^i$ and $\phi^i$ will repeat for $M$ iterations to obtain M workers(($\overline{\theta}^1, \phi^1$),($\overline{\theta}^2,\phi^2$),...,($\overline{\theta}^M,\phi^M$)), which also can be done in parallel.
In meta-training, the scores of rollouts by each $\overline{\theta}^i$ and $\phi^i$ after adaptation training will be used by the fitness functions $F_{actor}(\theta^i_1, \theta^i_{2},...,\theta^i_{K})$ and $F_{critic}(\phi^i)$
with its corresponding parameters to update the meta-distribution. The outer meta-training loop and the inner adaptation loop will continue until a good meta-distribution is found.

\subsection{Fast Adaptation and Sampling}
The basic idea of fast adaptation in adaptation training is to adapt to a new task with a small amount of experience, 
which means that the agent needs efficient exploration and high sample-efficiency. 
Off-policy methods usually have comparatively higher sample-efficiency than on-policy methods. 
Also, deterministic policy gradient methods have higher sample-efficiency than stochastic policy gradient methods. 
Both of these greatly improve the convergence of adaptation training. 
Parameter space noise is the key for consistent and efficient exploration in new tasks. 

In our algorithms, the samples for i-th worker are sampled by $K$ perturbation models($\theta^i_1$,$\theta^i_2$...$\theta^i_K$). These $K$ models follow the original distribution $\mathcal{N}(\mu_a,\,\sigma_a^{2})$. And the means of the $K$ perturbation parameters construct the initial parameters $\overline{\theta}^i$ in the beginning of adaptation training, which subjects the distribution of $\mathcal{N}(\mu_a,\,\frac{\sigma^{2}_a}{\mathcal{K}})$.
The $K$ samples have higher deviations for better exploration while the initial policy with lower deviation can make the learning more stable. 

\subsection{Learning Distribution of Exploration Parameters by Evolution Strategies}
We use Deep Deterministic Policy Gradient (DDPG) to perform adaptive training, hence we need to co-evolve actors and critics. 
In order to minimize the squared error between the critics after adaptive training and the Q value, 
the critics' fitness function is the negative of the squared error between the adapted critic and the test set(update).
\begin{equation}
\label{critic_meta_params}
\begin{aligned}
\hat\mu_{critic} , \hat\sigma_{critic} & =  \mathop{\arg\max}_{\mu_c\,\sigma_c}(\mathcal{J}_{critic}(\mu_c,\sigma_c))\\
\mathcal{J}_{critic}(\mu_c,\sigma_c) &= \int F_{critic}(\phi)\mathcal{N}{(\phi | \mu_c,\sigma_c)}d\phi
\end{aligned}
\end{equation}
The actors' fitness function $F(\theta_{1},\theta_{2},...,\theta_{K})$ is the score of rollout by the adaptive trained paramters $\theta'$. 
The goal of evolution strategies for actors is to find a Gaussian distribution $\mathcal{N}({\mu_a},\,{\sigma_a}^{2})$, 
so that sampling $\theta_{1}$,$\theta_{2}$...$\theta_{K}$ from it can maximize the expectation of $F(\theta_{1},\theta_{2},...,\theta_{K})$. 
Since $\theta_{1},\theta_{2}...\theta_{K}$ are independent, their joint distribution is ${\prod_{1\le i \le K} \mathcal{N}(\theta_i | \mu_a,\sigma_a^{2})}$. 
The goal of the whole algorithm becomes finding $\hat\mu_{actor}, \hat\sigma_{actor}$:
\begin{equation}
\label{actor_meta_params}
\begin{aligned}
\hat\mu_{actor} , \hat\sigma_{actor} & =  \mathop{\arg\max}_{\mu_a\,\sigma_a}(\mathcal{J}_{actor}(\mu_a,\sigma_a))\\
\mathcal{J}_{actor}(\mu_a ,\sigma_a) &= \int...\int F_{actor}(\theta_1...\theta_K)\\& {\prod_{1\le j \le K} \mathcal{N}(\theta_j | \mu_a,\sigma_a)}d\theta_1...d\theta_K
\end{aligned}
\end{equation}
The gradients of $\mu$ and $\sigma$ in adaptation training are defined as below. Note that both gradients are independent of the fitness functions used in meta-training.
\begin{equation}
\begin{aligned}
\nabla \mathcal{J}_{critic}(\mu_c,\sigma_c) &= \int F_{critic}(\phi) \nabla \mathcal{N}(\phi| \mu_c,\sigma_c)d\phi\\
&= \int F_{critic}(\phi) \nabla \mathcal{N}(\phi | \mu_c,\sigma_c) \frac{\mathcal{N}(\phi | \mu_c,\sigma_c)}{\mathcal{N}(\phi | \mu_c,\sigma_c)}d\phi\\
&=  \int F_{critic}(\phi) \nabla log { \mathcal{N}(\phi| \mu_c,\sigma_c)}\mathcal{N}(\phi | \mu_c,\sigma_c)d\phi\\
&= \frac{1}{M}\sum_{1 \le i \le M}F_{critic}(\phi_{i}) \nabla log{ \mathcal{N}(\phi_{i} | \mu_c,\sigma_c)}
\end{aligned}
\end{equation}
\begin{equation}
\begin{aligned}
    \nabla \mathcal{J}_{actor}(\mu_a,\sigma_a) &= \int...\int F_{actor}(\theta_1...\theta_K) \nabla {\prod_{1\le j \le K} \mathcal{N}(\theta_j | \mu_a,\sigma_a)}d\theta_1...d\theta_K\\
    &= \int...\int F_{actor}(\theta_1...\theta_K) \nabla {\prod_{1\le j \le K} \mathcal{N}(\theta_j | \mu_a,\sigma_a)} \\
    &\frac{\prod_{1\le j \le K} \mathcal{N}(\theta_j | \mu_a,\sigma_a)}{\prod_{1\le j \le K} \mathcal{N}(\theta_j | \mu_a,\sigma_a)}d\theta_1...d\theta_K\\
    &=  \int...\int F_{actor}(\theta_1...\theta_K) \nabla \sum_{1\le i \le K} log { \mathcal{N}(\theta_i | \mu_a,\sigma_a)} \\
    &{\prod_{1\le j \le K} \mathcal{N}(\theta_j | \mu_a,\sigma_a)}d\theta_1...d\theta_K\\
    &= \frac{1}{M}\sum_{1 \le i \le M}F_{actor}(\theta^{i}_{1}...\theta^{i}_{K})\\& \nabla \sum_{1\le j \le K} log { \mathcal{N}(\theta^{i}_{j} | \mu_a,\sigma_a)}
\end{aligned}
\end{equation}
Since the meta-distribution follows a Gaussian distribution, the gradients of $\mu , \sigma$ can be derived as:
\begin{equation}
\begin{aligned}
\label{training_equations}
\nabla_{\mu_a} \mathcal{J}_{actor}(\theta) &= \frac{1}{M}\sum_{1 \le i \le M}F_{actor}(\theta^i_{1}...\theta^i_{K})\sum_{1\le j \le K}\frac{\theta^i_{j}-\mu_a}{\sigma_a^2}\\
\nabla_{\sigma_a} \mathcal{J}_{actor}(\theta) &= \frac{1}{M}\sum_{1 \le i \le M}F_{actor}(\theta^i_{1}...\theta^i_{K})\sum_{1\le j \le K}\frac{(\theta^i_{j}-\mu_a)^2-\sigma_a^2}{\sigma_a^3}\\
\nabla_{\mu_c} \mathcal{J}_{critic}(\phi) &= \frac{1}{M}\sum_{1 \le i \le M}F_{critic}(\phi_i)\frac{\phi_{i}-\mu_c}{\sigma_c^2}\\
\nabla_{\sigma_c} \mathcal{J}_{critic}(\phi) &= \frac{1}{M}\sum_{1 \le i  \le M}F_{critic}(\phi_i)\frac{(\phi_{i}-\mu_c)^2-\sigma_c^2}{\sigma_c^3}
\end{aligned}
\end{equation}
\subsection{Meta-distribution Learning Algorithms}
\begin{algorithm}[htb]
\setstretch{1.35}
\caption{}
\label{alg:Framwork}
\begin{algorithmic}
\STATE Initialize the parameters of distributions  $P_{actor}(\theta|\mu_a,\sigma_a)$ and $P_{critic}(\phi|\mu_c,\sigma_c)$\
\WHILE{not done}
\STATE Sample a mini-batch of tasks $\mathcal{T}_t$ from $p(\mathcal{T})$\
\FOR{i in M}
\STATE sample K actor parameters $\theta^i_1,\theta^i_2,...,\theta^i_K$ from $P_{actor}(\theta|\mu_a,\sigma_a)$ and critic parameter $\phi^i$ from $P_{critic}(\phi|\mu_c,\sigma_c)$\
\FOR{each task $t \in \mathcal{T}_t$}
\STATE get K trajectories based on K actor parameters
\STATE initialize actor parameter $\overline{\theta^i}$ based on K actor parameters $\theta^i_1,\theta^i_2,...,\theta^i_K$\
\STATE use K trajectories and critic to adaptively train $\overline{\theta^i}$ and $\phi^i$
\STATE get adapted parameters $\overline{\theta^i}'$ and ${\phi^i}'$ to sample fitness of actor and critic in task $t$
\ENDFOR
\ENDFOR
\STATE update $P_{actor}(\theta|\mu_a,\sigma_a)$ and $P_{critic}(\phi|\mu_c,\sigma_c)$ using Equation 10\

\ENDWHILE
\end{algorithmic}
\end{algorithm}

\section{Experiment}
Meta-learning in reinforcement learning is analogous to few-shot learning in supervised learning. After training on a collection of tasks, the agent should be able to learn a new task with only a small amount of further training. That is to say we are given a distribution that encompasses both the collection of tasks in the training set and the new task in the test set. In our algorithm, the ideal agent has the optimal starting parameters from which to start exploring efficiently when learning a new task.



Broadly speaking, a new task might consist of achieving a new goal, operating under a new environment or a different transition distribution, yet all tasks must be from the same task distribution.
In our experiment, we use the $rllab$ benchmark environment \cite{Duan2016Benchmarking} which is a simulated continuous control environment. More specifically, we used a modified version designed for Model-Agnostic Meta-Learning. \cite{Finn2017Model}

\begin{figure*}[!th]
	\centering 
	\includegraphics[width=0.4\textwidth]{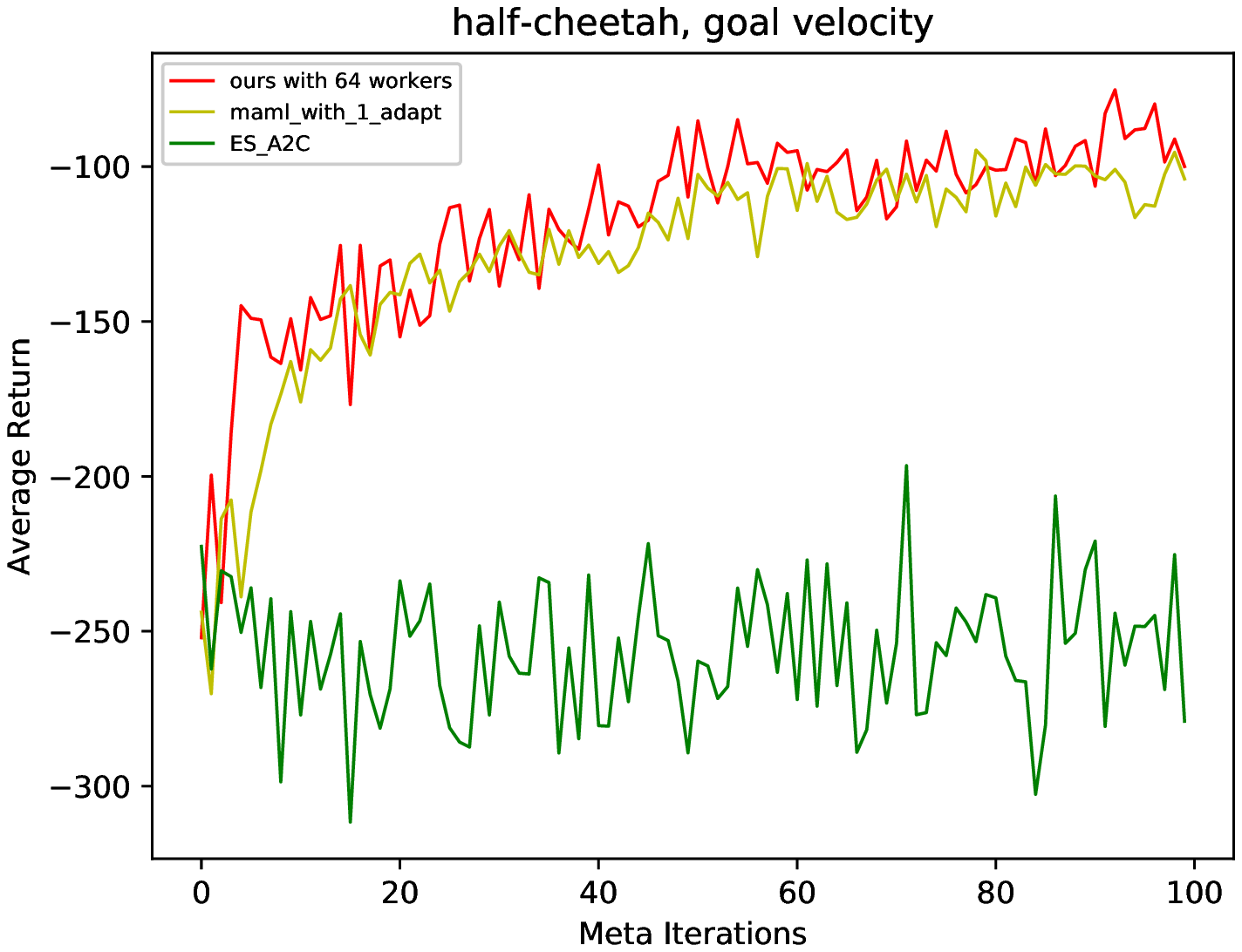} 
	\includegraphics[width=0.4\textwidth]{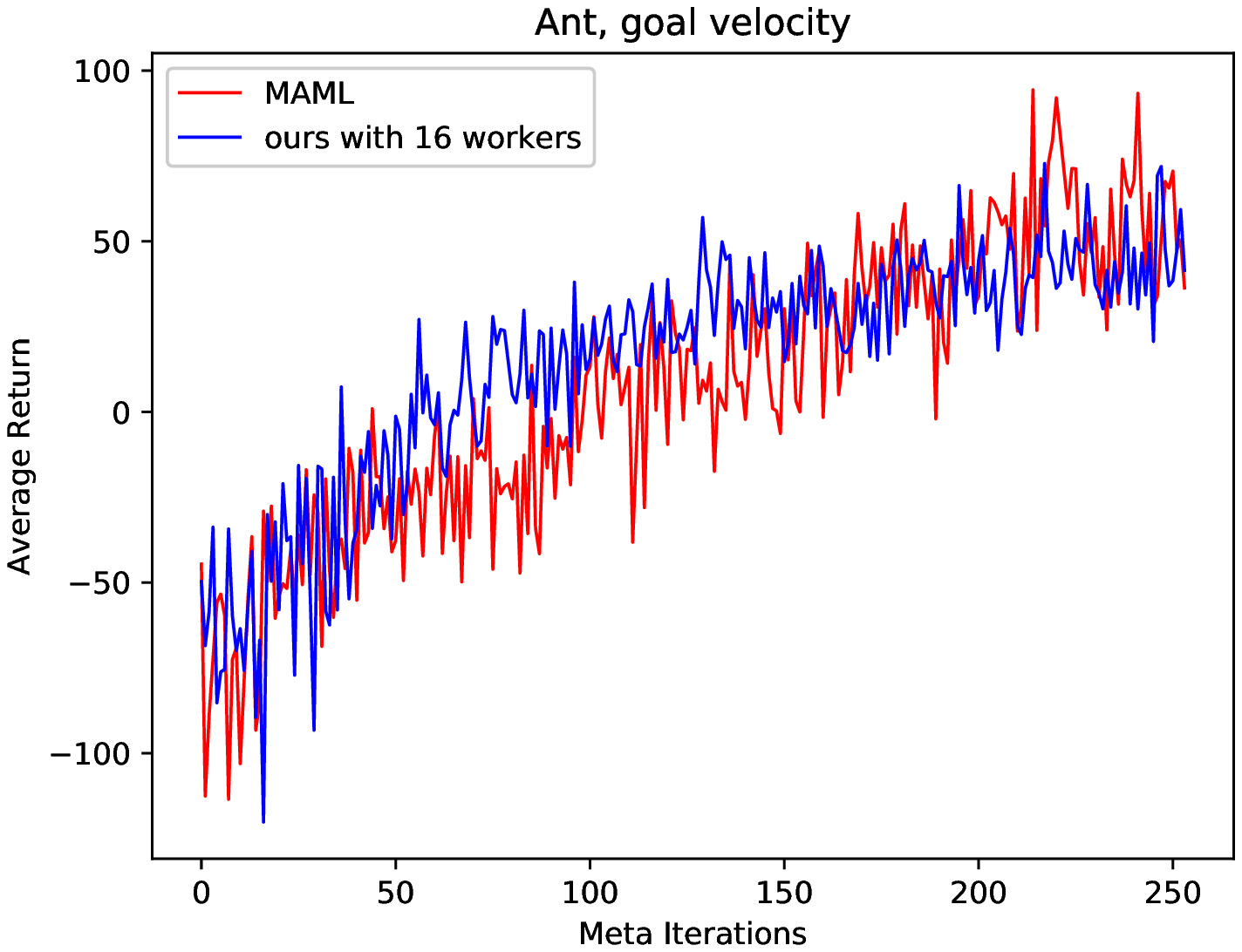} 
	\includegraphics[width=0.4\textwidth]{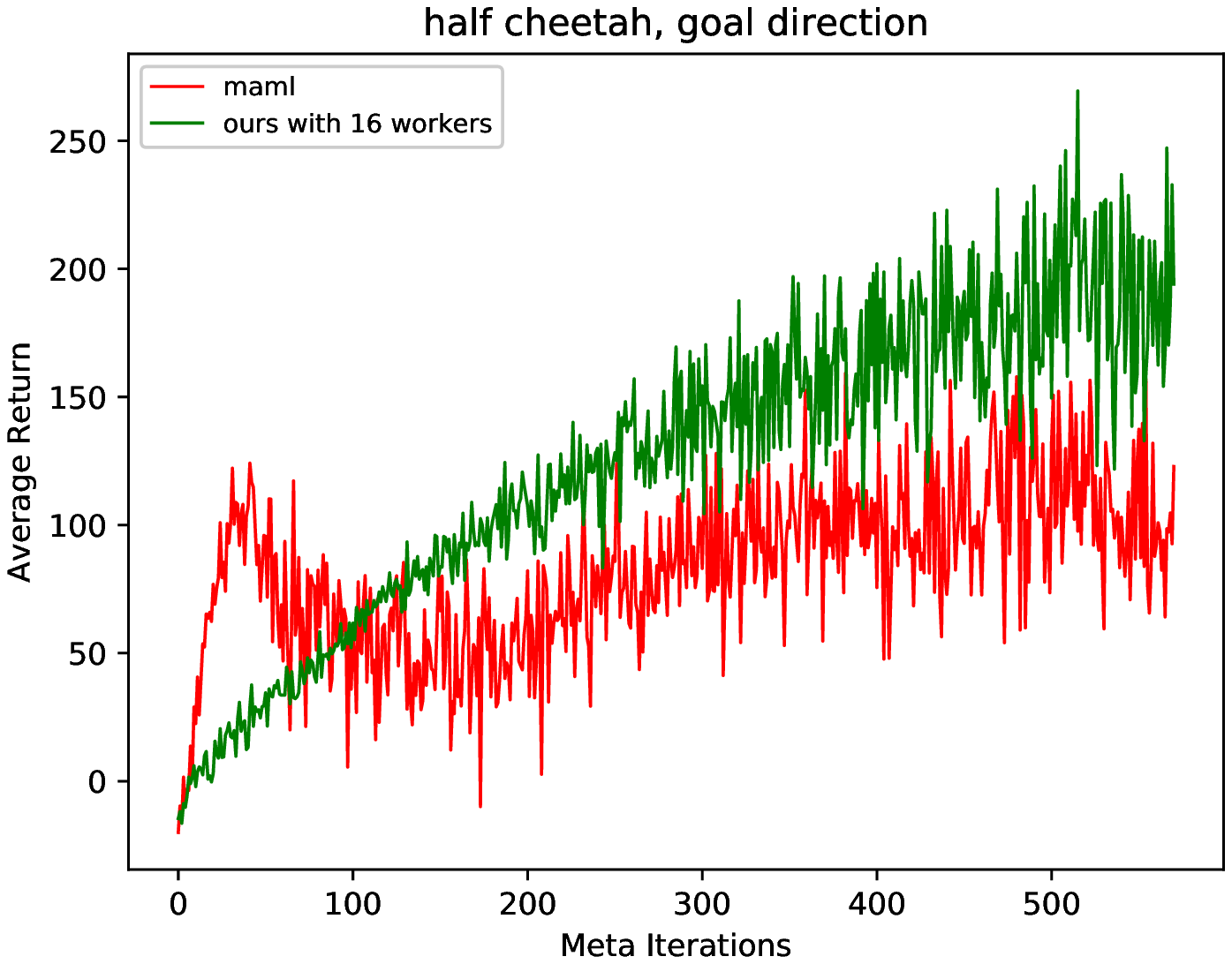} 
	\includegraphics[width=0.4\textwidth]{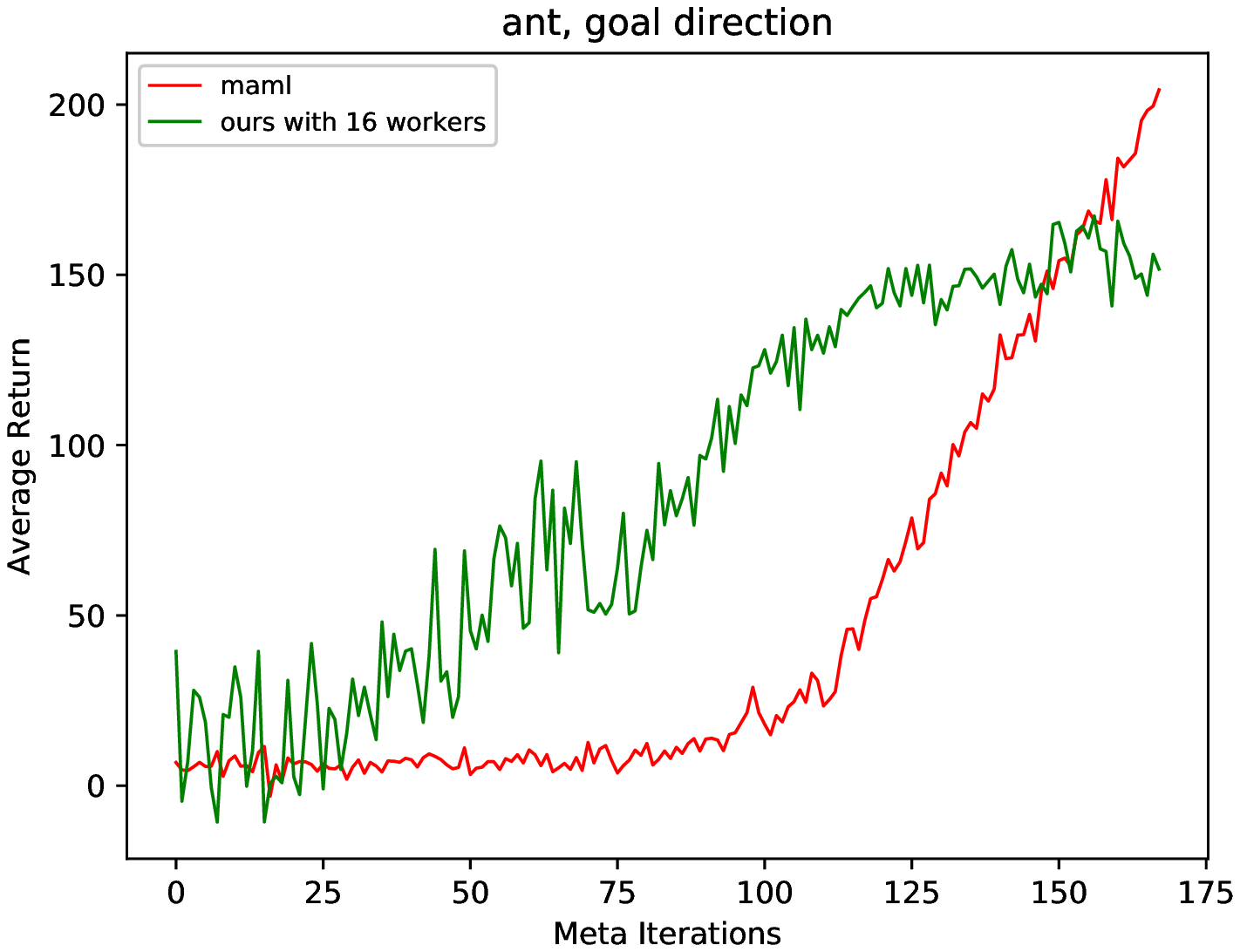} 
	\caption{Our experimental results of the half-cheetah and ant robot for goal velocity and goal direction compared with MAML.}
\end{figure*}

\subsection{Environments}
The high-dimensional control tasks in the MuJoCo simulator \cite{Todorov2012MuJoCo} can be divided into four categories, including controlling two simulated robots to achieve two kinds of goals. The simulated robots in our tasks are a planar cheetah and a 3D quadruped (the `ant'). The specific goals are to control the robots to run at a particular velocity or in a particular direction, and the corresponding target values of goals are not used as an input to the robot (i.e. the robot needs to explore or find out the target value using experience sampled during adaptation training). When needing to run at a particular velocity, the reward is the sum of the negative absolute value between the velocity of the agent and a goal, sampled uniformly between 0.0 and 2.0 for the cheetah, and between 0.0 and 3.0 for the ant. When the goal is to follow a particular direction, the reward is defined as the sum of the magnitude of the velocity in the corresponding direction (forward or backward). 

Our environment setting is exactly the same as described in MAML. The horizon for each task is H = 200. In every iteration, the amount of experience used for adaptation training for a specific task for each worker in our algorithm is the same as in MAML, which is 20 trajectories per iteration for all problems except the ant forward/backward task, which uses 40 trajectories.

\subsection{Implementation and Details}

In all tasks, the specific model for each worker is an actor neural network and a critic neural network. The actor network's input is an observation of the environment, and its output is an action. The network has two hidden layers of size 100 with ReLU nonlinearities. Xavier's random weight initialization is used for each neuron. The critic network evaluating state-action values has the same structure as the actor, but the input of the second hidden layer is the output of the first hidden layer concatenated with the action. Besides these networks describing the mean of the Gaussian distribution (Eq. \ref{critic_meta_params}, Eq. \ref{actor_meta_params}), two extra sets of variables describing the corresponding standard deviation are also used to help training and exploration. Hence, we use four Stochastic Gradient Descent (SGD) optimizers to train the related variables. We find it worth noting that the Adam optimizer(\cite{Adam}) does not help for the training of evolution strategies. The specific meta-training steps are performed according to the formula in Eq. \ref{training_equations}.

During every iteration, for each worker, we use some random seeds (20 or 40 depending on the experiment) to perturb the meta-network to produce corresponding exploration actor networks($\theta^{i}_1, \theta^{i}_2,...,\theta^{i}_k$). These exploration actor networks will sample trajectories to build a temporal replay buffer for the central network($\overline{\theta}^i$), whose parameters are the means of parameters of the exploration networks. The central network is trained and then performs a rollout once to calculate the fitness of this worker. After using all workers to evolve these seeds, a new meta network will be broadcast to each worker. The above procedure is repeated until an appropriate meta-policy has been achieved.
\subsection{Experimental Results}\label{experimental_results}

Under the same amount of computation for every worker, our algorithm can also achieve the better performance than the two-order adaptive training method mentioned in MAML\cite{Finn2017Model} for tasks of goal velocity. As illustrated in the top left of figure 1, especially when using 64 workers, our algorithm can converge faster and achieve better results than MAML. And we predict that when using more workers, much more improvement will be attained. In addition, We have also tried the methods in \cite{DBLP:journals/corr/abs-1806-07917}, which combines the A2C and ES optimization. But it fails when applying to the high dimension environments as figure 1 shows. And for the task half-cheetah of goal direction, our algorithm can achieve to 400 average return after about 3000 iterations. For the task of goal direction for ant, although the convergence is fast, we find that the performance is not as prominent as the tasks of goal velocity which MAML can converge to 480 average return after about 600 iteration, probably because a set of perfect hyper-parameters is difficult to search. And we prepare to solve this problem in the future.


\begin{figure*}[!th]
	\centering 
	\includegraphics[width=0.4\textwidth]{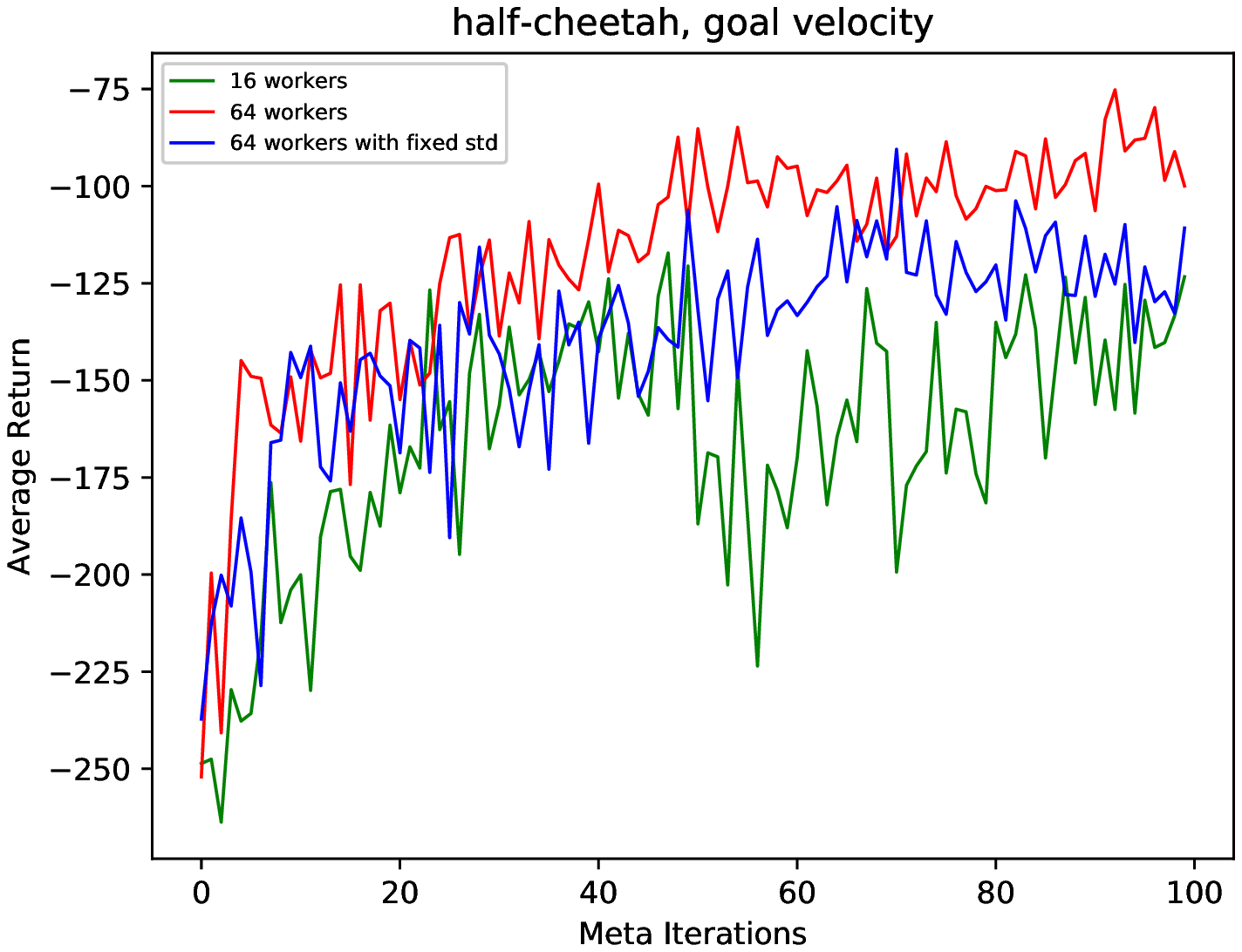} 
	\includegraphics[width=0.4\textwidth]{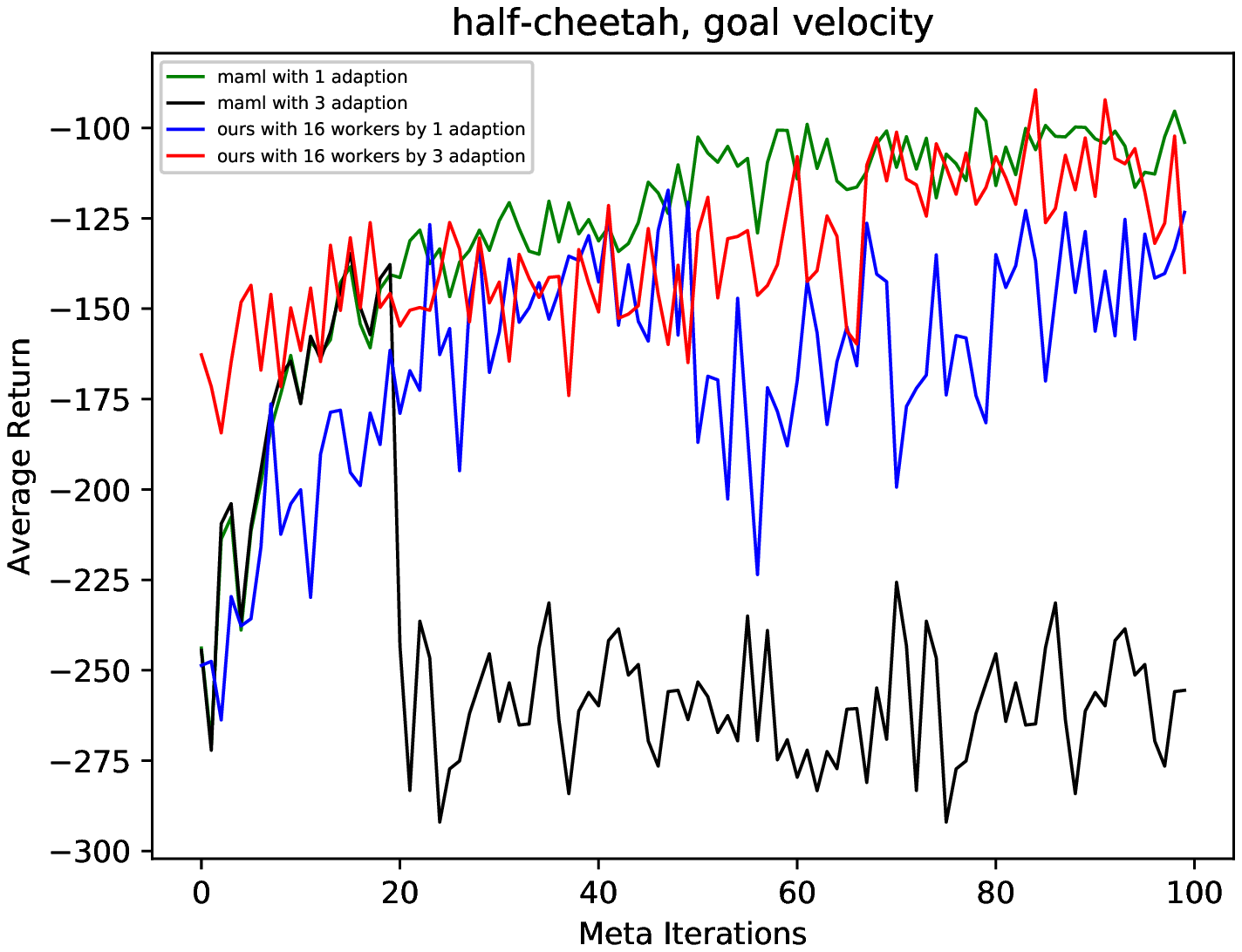} 
	\caption{left: Compare the performance of high parallel and whether the standard deviation is fixed; right: different gradient update times in the adaptation training for the MAML and our algorithm.}
\end{figure*}

In addition to comparison with MAML, we also verify the following points: (1) As described in \cite{Salimans2017Evolution}, highly parallel, evolution strategy-based optimization algorithms can achieve linear speedups even when using multiple workers, so we show speedups can also apply to our algorithm; (2) Due to the off-policy algorithm \cite{DDPG}, our algorithm can achieve better results when performing several adaptive training steps repeatedly using the experience sampled by the exploration networks.(3) The standard deviation of the Gaussian distribution we are searching is helpful for workers' exploration and training. Using a trainable standard deviation instead of a fixed value enables agents to adjust the extent of exploration according to the task in order that more effective experience can be sampled during adaptation training.  We address these points separately below.

\begin{enumerate}
	\item \emph{Speedups By High Parallel}:As mentioned in \cite{Salimans2017Evolution}, an ES optimizer is particular amenable to parallelization when using random seeds, because it only requires infrequent communication after complete episodes. Here, we compare the performance between 16 workers and 64 workers. As illustrated in the left of Figure 2, for the given task, a greater number of workers can bring about a greater result more quickly and more stably.
			
	\item \emph{Data Efficiency}: Because of ignoring many higher-order derivatives that MAML needs to consider during meta-training, we predict that MAML may not work if executing several gradients update during adaptation training. However, ES, being a black box optimization algorithm, need not consider too much. We attempt to conduct the half-cheetah velocity experiment to compare the performance between MAML and our algorithm; we run MAML twice, once with 3 gradient updates in the adaptation training loop and another time with only 1 gradient update per loop. Results in the right of figure 2 show that the 3-time update in MAML fails to learn, while bringing better performance in our algorithm.

    \item \emph{Trainable Standard Deviation}: The standard deviation of the Gaussian distribution we are searching is helpful for workers' exploration and training. Using trainable standard deviations instead of fixed values enables agents to adjust the extent of exploration according to the task in order that more effective experience can be sampled during adaptation training. The left of figure 2 shows that by using a flexible standard deviation under identical hyper-parameters, a higher final score and more stable training can be achieved.
\end{enumerate}

\section{Discussion}

In this work, we introduce a method based on evolution strategies, which is comparable to gradient-based meta-learning methods in reinforcement learning. And our contributions are as follows.
\begin{enumerate}
\item \emph{}Our proposed method does not need higher-order gradients in multi-step task adaptation training and can perform multiple gradient updates in adaptation training, so it has more potential to learn well on more complex tasks, like playing different levels from a single video game. 

\item \emph{}We propose a framework of meta reinforcement learning, which using evolution strategy to learn a meta model in the outer loop and using any reinforcement algorithm(DQN,A3C,TRPO,PPO,DDPG) to do adaption for specific task because of the zero-order property of evolution strategy. 

\item \emph{}The evolution strategy can learn a distribution of exploration parameters and obtain the initial parameters in a simple way and use noise perturbation points to approximate a mean value, as opposed to the standard method of creating noise from a mean. so we can evolve our policy distributions without using higher-order gradients.

\item \emph{}The evolution strategy as a black box method can be parallelized on a large scale, which is easy to implement, and our work also proofs that it can work well even for complex meta learning tasks, which shows the enormous potential of the evolutionary strategy optimization algorithm.
\end{enumerate}

\subsection{Future Work}\label{future}
In our algorithm implementation, there are so many hyper-parameters to be fine-tuned, such as the learning rate in the meta-training stage and adaption stage for actor and critic network. And it is difficult to find a suitable set of parameters in some tasks(for example, the task of ant goal velocity). We want to optimize the algorithm to reduce the amount of hyper-parameters, or introduce some algorithms that can automatically find the hyper-parameters, such as PBT(\cite{PBT}).
On the other hand, during every meta iteration, the initial parameters for adaption is the means of K perturbed model parameters, which may bring a little randomness and instability. 
In future research, we plan to use more flexible ways to obtain the initial parameters of new tasks and more types of noise distributions to improve the stability of our algorithm.

\bibliographystyle{aaai}
\bibliography{ref}
\end{document}